\pdfoutput=1
\documentclass[11pt]{article}

\usepackage[]{acl}

\usepackage{times}
\usepackage{latexsym}

\usepackage[T1]{fontenc}
\usepackage[utf8]{inputenc}

\usepackage{microtype}

\usepackage{amsmath}

\usepackage{graphicx}
\usepackage{caption}
\usepackage{subcaption}
\usepackage{tikz-dependency}

\title{CALaMo: a Constructionist Assessment of Language Models}

\author{Ludovica Pannitto\\
  CIMeC\\
  University of Trento\\
  \texttt{ludovica.pannitto@unitn.it} \\\And
  Aurélie Herbelot\\
  CIMeC/DISI\\
  University of Trento\\
  \texttt{aurelie.herbelot@unitn.it} \\}

\begin{document}
\maketitle
\begin{abstract}
This paper presents a novel framework for evaluating Neural Language Models' linguistic abilities using a constructionist approach. Not only is the usage-based model in line with the underlying stochastic philosophy of neural architectures, but it also allows the linguist to keep meaning as a determinant factor in the analysis. We outline the framework and present two possible scenarios for its application.
\end{abstract}

\section{Introduction}

Over the years, linguists have given a lot of thought to what language \textit{is}, and how it can be best formally described. 
%Their object of study is known to be extremely multifaceted: 
Different approaches with sometimes contradictory aims have produced an extremely rich array of conceptual tools to describe linguistic phenomena. Such tools play diverse roles in explaining the phylogenetic, ontogenetic or historical-cultural facets of language and are often heavily interlaced with one another. In this research landscape, computational modelling has largely been used to simulate and investigate speaker behaviour at various levels of granularity. A specific area within the computational community is known as \textit{(Neural) Language Modelling}, which aims at reproducing linguistic surface structure by means of (pseudo)-probabilistic models. Neural architectures have played a special role in this subfield of research, due to their flexibility. %Computational modelling is also more and more crucial as a reverse-engineering approach to tackling known questions in cognitive science or psycholinguistics~\citep{Dupoux2018-cognitive}.

The extreme complexity of theoretical tools found in linguistics %that is, in the study of language as a human means of expression, 
 gets cut down by order of magnitudes when it comes to the analysis of language processing using computational modelling. %In the area of NLM, in fact, whether systems acquire any sort of linguistic knowledge remains one of the biggest conundrums of the field. 
 For instance, when the term \textit{language} is mentioned in relation to Artificial Neural Networks, it seems that the word is often used as a mere synonym of \textit{grammar}: while it is clear from a broader theoretical perspective that the two objects do not overlap, the distinction gets blurred in many computational studies. That is, assumptions which would be clearly stated in theoretical linguistics (e.g. how grammatical abstraction fits into the concept of \textit{language}), are not explicitly discussed by computational studies: it is often the case that a specific set of choices concerning the description of language are taken as default.
Most current work also seems to implicitly make a number of assumptions about what kind of grammar is supposed to emerge from neural language models (henceforth, NLMs), and this underlying choice is often echoed in the most common evaluation settings and in the conclusions that are being drawn from such experiments.
Most of these default assumptions are inherited from the nativist Chomskian tradition and the Universal Grammar (UG) framework~\citep{chomsky1986knowledge,smith2016chomsky}, which has pervaded a lot of the computational work on grammar, and continues to do so in the recent literature on neural models. 
%In our literature review (\S\ref{ch:literature}), we will talk in more details about the specific postulates that have been integrated into current frameworks, and whether this integration was warranted, given the architecture and learning behaviour of neural models.

Ironically, the nativist assumptions that permeate the mainstream computational methodology are at odds with the very nature of the models created by the field. Neural models are essentially based on pattern learning and are completely agnostic about the nature of the data they are made to process. The idea that language can be abstracted from a general purpose statistical mechanism is more akin to usage-based (henceforth, UB) approaches~\cite{barlow2000usage,goldberg2003constructions,tomasello2003constructing}, and NLMs would provide a much more natural testbed for that theoretical strand.
%In one of the foundational works of the UB theorization, for instance, \cite{Tomasello2022-cultural} highlights how the acquisition of language can actually be part of a wider adaptation for cultural learning in general, and that pre-existing mechanisms such as schematization, categorization, statistical learning and analogy-making, already present in primates, determined the grammaticalization of linguistic structures and are enough to do so on the ontogenetic level, too.
In the cognitive and UB accounts, the exploitation of predictability during language development (and again we refer to development at all the three tiers of philogeny, ontogeny and cultural evolution) is the root of a number of fundamental mechanisms such as schematization, entrenchment and distributional analysis~\citep{Lewkowicz2018-learning}.
In the light of these processes, language, seen as a structured inventory of constructions, gets build through generations~\citep{cornish2017sequence} and throughout a speaker's lifetime: shared linguistic material among utterances, such as morphological markers for instance, enable the identification of particular patterns or constructions as units bearing meaning~\citep{Croft2001-radical}.

The perceived gap between the nativist and non-nativist traditions with respect to computational modelling probably stems from historical factors. %the fact that different theories have emerged from different communities, and only some of them have co-evolved with the computational modelling community: among these, 
The Chomskian school and its formal approach offered a definition of language that, in the past, could easily be interpreted and implemented by emergent computational approaches. But there is no reason for this bias to perdure. In this paper, we argue in favour of a usage-based framework to analyse language acquisition in ANNs. We first point out the aspects of nativist theories that have so far influenced the evaluation of NLMs (\S\ref{sec:nativist}). We then introduce a framework for a quantitative and qualitative analysis of NLMs linguistic abilities within the constructionist perspective (\S\ref{sec:formalization}). We finally show some preliminary analyses performed with the proposed formalization (\S\ref{sec:experiments}).

%Hence, it is within the computational community that it has left most traces.

%The purpose of this work is to look at the current mainstream training and evaluation methodology for neural language models, taking into explicit consideration the linguistic assumptions made along the way. Throughout this thesis, we will take the idea of a `model' seriously, i.e. we regard the computational framework as a way to simulate some part of reality in a vacuum, making explicit simplifying assumptions in the process. We  argue for a tighter integration between the description of such a model and the linguistic theory that it is supposed to simulate. That is, we would want to see how the working parts of a model encode specific theoretical statements, and most importantly, we would like to know where the model simplifies and where it directly contradicts the theory. To that end, we will propose an improved methodology to tie up computational modelling and linguistic frameworks, and we will test this methodology in two actual case studies relating to the grammatical abilities of Recurrent Neural Networks.

\section{Nativist vs. non-nativist approaches to language acquisition}
\label{sec:nativist}

All theories of language use and development recognize that at the root of human linguistic ability is the capacity to handle symbolic structures. But they disagree on the specific content of speakers' linguistic knowledge, the mode of acquisition of such content, and the extent to which linguistic productivity is affected by this stored knowledge~\citep{bannard2009modeling}. Theories diverge with respect to three aspects: input, stability and systematicity. The perspective taken on each of these aspects has consequences for the conclusions drawn from NLMs' responses to the evaluation setting.  In the following, we consider each aspect in turn and specifically highlight how the evaluation of computational models becomes biased due to a lack of explicitness in relating experimental and theoretical aspects of the research question. %Such bias is evident at the quantitative level (i.e. the chosen evaluation measures may push interpretation in a certain direction), but also in the analysis of results, where discussions are often marred with implicit postulates.

\paragraph{Input.} One of the main arguments introduced by nativist frameworks is the \textit{poverty of the stimulus}: the input children are exposed to is underdetermined and does not explain acquisitional generalizations observed in learners~\citep{Crain2001-nature}. Such theories assume that children navigate a hypotheses space defined by innate constraints~\citep{Eisenbeis2009-generative}. Constructionist approaches, instead, posit that language emerges from the input through domain-general mechanisms: this implies that the input is shaped and skewed in a specific way in order to enhance learnability~\citep{Boyd2009-input}. A well established line of research has shown how children are  proficient statistical learners \citep{gomez2000infant,romberg2010statistical,christiansen2019implicit}. The emergence of language-like structure from purely linear signal has also been shown in recent experiments such as ~\cite{cornish2017sequence}, which demonstrated how important aspects of the sequential structure of language may derive from adaptations to the cognitive limitations of human learners and users~\citep{Christiansen2016-now}.
The crucial difference between the nativist and the non-nativist approach here is how strict the relation between the received input and the acquired linguistic structure is: if we commit to a view in which the input only serves as a trigger of an almost pre-determined cognitive structure, we are naturally driving our attention far from the features of the input and primarily to the features of the structure. On the other hand, deriving the linguistic structure from the input structure itself requires investigating the two aspects together. So far, most studies on NLMs have disregarded the effect of the input on experimental results~\cite{pannitto2022can}.

\paragraph{Stability.} The \textit{continuity assumption} was first introduced by \citet{Pinker1984-language} in order to reconcile aspects of developmental language with the generative framework. It posits that the differences between adult and children linguistic structures is negligible and merely due to performance factors. In contrast, what we can refer to as the \textit{developmental hypothesis} claims that the mechanisms underlying acquisition remain the same throughout a life-long acquisition process, but the structures and abstractions they generate evolve over time. UB models also put emphasis on the linear and time-dependent nature of the linguistic signal~\citep{Christiansen2016-now,cornish2017sequence}. According to the UB account, generalizations appear gradually, as productivity emerges from item-specific knowledge~\citep{Bannard2009-modeling}.
%the evidence about child's own knowledge of grammatical structure is in fact contradictory~\citep{Dittmar2008-young,Gertner2006-learning}.

Another aspect of stability is inter-speaker differences. UG posits that all speakers eventually converge to the same grammar~\citep{Lidz2009-constructions,Crain2009-capturing}. Individual differences have however been found in almost every area of grammar, depending on a variety of factors including environmental ones~\citep{Street2010-more}. The `sameness' assumption pervades the computational linguistics literature, where evaluation is performed according to a single `gold standard' per task. For traditional tasks such as sentiment analysis or word similarity ratings, the annotations of human subjects are averaged, and the system is evaluated against the average. For language modeling, model perplexity is computed with respect to the statistical features of a large corpus, which aggregates the writing styles and linguistic habits of thousands of speakers. While this state of affairs has started to be criticised by various researchers, it remains for now the status quo.
When considering language development as a speaker-dependent process, strongly affected by the nature of the input, an evaluation based on an `average speaker' becomes truly unsatisfactory. We cannot assume the existence of a ground truth, and must rely on softer evaluation measures: it is clear that the linguistic behaviours of different speakers must overlap sufficiently to allow for communication, but that we also want to observe in the output of the network the kind of variability that is seen in humans. 

%It is again a matter of framing: as \cite{Dabrowska2015-what} points out, individual differences are not in principle incompatible with innateness, as they could be as well due to inheritance factors. We do not, however, necessarily want to assume continuity and stability when it comes to the description of grammar itself.
%Individual variation is similarly suppressed by the idealized \textit{native speaker} posited by UG: the almost exclusive focus on linguistic competence has automatically implied the exclusion of non-native proficiency from the scope of investigation. As \cite{Radwanska-Williams2008-native} point out, the idea of a linguistic community as an homogeneous system stems from the structuralist tradition, where language was described synchronically as the result of the aggregation of different realizations in the speech community. The Chomskian tradition projects such homogeneity on the single speaker: what was originally conceptualized as a social dimension has thus become an individual feature.

\paragraph{Systematicity.} The ability to understand and generate an unbounded number of novel sentences, using finite means, is  considered one of the hallmarks of our language faculty. The boundaries of this systematicity remain however largely unclear: provided that we agree on what the finite means at our disposal are, not all the possibilities are actually realised by speakers and not all realised possibilities share the same cognitive or linguistic status.

One way to look at systematicity is that of compositionality, for which the most widely known version is probably due to~\citet{Katz1963-structure}, that port Chomsky's innateness theory to semantics: a set of rules or constraints is needed in order to systematically build the meaning of sentences by integrating meaning of words. 
Even the Montagovian formal approach to compositionality~\citep{Montague1970-english} relies on Chomskian-derived ideas of a stable lexicon that stores meanings, and the existence of a set of precise interpretation rules that allow for those meanings to be mixed and modulated \textit{through} the filter of syntax.
The core of both visions is still very much syntax-centered (to which semantics has to be isomorphic) and very little space is left for indeterminacy, negotiation between speakers and other aspects related to the interactive and communicative nature of language (different individuals can retain in fact quite different concepts associated to the same lexical label for instance, \citealp{Labov1973-boundaries}).
In a nutshell, if we see systematicity from the standpoint of compositionality, the quasi-regularities of linguistic structure represents a major hurdle to surpass. 

Quasi-regularity is instead the engine of productivity, as in the ability of speakers to use all the available linguistic means to cue the intended meaning. 
Just like compositionality, productivity deals with the domain in which a grammatical pattern can be employed in a linguistic context without losing interpretability, and it deals with what is actually possible in the language and where to draw the boundaries of acceptability.
The shift has not just been syntactic: in the formal representation of these two aspects of systematicity in semantics, for instance, composition-oriented~\citep{Katz1963-structure} or productivity-oriented~\citep{Fillmore1976-frame} theories have conceptualized the idea of selectional constraints differently. 
%In~\citet{Katz1963-structure} for instance the idea is that formal constraints are hard-coded in the lexicon, in order to regulate ambiguity and semantic acceptability. Later on, for instance in \citet{Fillmore1976-frame}'s model, selectional constraints are relaxed to selectional preferences, making space for individual manipulation in rules and boundaries.

Knowledge on systematicity is in both cases considered as implicit knowledge that the speaker has about their language.
Nativist approaches have however primarily dealt with compositionality, and so are NLMs often evaluated: given grammar rules and lexicon, what are the computational mechanisms that allow them to combine?
UB theories, on the other hand, have primarily been dealing with productivity: how far can meaning boundaries be forced? What are the mechanisms that allow for linguistic creativity?
This of course entails, in the UB community, a relation to surface properties of the input as well: \citet{Croft2004-cognitive}, for instance, note how the maximally schematic constructions, such as \texttt{sbj verb obj}, are also the most productive ones, and that this has a relation to their frequency too, both as a type and for each of their instantiations.

\section{CALaMo}
\label{sec:formalization}

In our proposed methodology, CALaMo (Constructionist Assessment of Language Models), we incorporate the UB perspective across all three aspects: input, stability and systematicity.

As far as \textit{input} is concerned, CALaMo differs from standard approaches by considering input data an important factor in determining the shape of the learner's grammatical knowledge. In traditional scenarios, the input only serves as a triggering factor and its features play little to no role in the analysis. From a UB perspective, instead, the relation between the abstract grammatical structure of the input and the acquired grammar, which then constrains the production of the learner, is strict.

Regarding \textit{stability}, depending on the view that is taken on the continuity hypothesis, we can see NLM's grammatical competence either as a binary or as a gradient property. In the first case, we test whether the network is able or not to handle some linguistic phenomenon, while in the second case, as advocated by CALaMo, we are interested in seeing how and why some linguistic aspect becomes more and more salient to the network during training.

The \textit{compositionality vs. productivity} perspectives, finally, entail a different organization of linguistic knowledge: the mainstream compositionality perspective tends to set meaning aside, and treat the lexicon as an organized repository of meanings (it makes sense therefore to test NLM's capabilities on semantically nonsensical sentences or to extend the known rules to completely unknown lexical items). In the productivity perspective, instead, meaning is intrinsically part of the process and is treated as a systematic aspect of grammar, too.

\subsection{Acquiring language}

When talking about NLMs and their linguistic capabilities, the issue of language acquisition ($A$) is often formalized as how much language $\Lambda$ can be learned by the (artificial) speaker, given a certain level of computational complexity $C$ by being exposed to a certain type of data $I$:
\begin{equation}
\label{eq1}
    A: C \times I \mapsto \Lambda
\end{equation} 

\noindent{}All the components of the equation have been central to the linguistic debate. However, starting from this basic formalization, we identify two major focus points that we specifically address in our framework. Firstly, the above formula describes acquisition as instantaneous, but it is actually better described as a process $A = (a_0, a_1, \cdots a_N)$ (\S\ref{steps}). From a cognitive perspective the process is fully continuous, while in the artificial scenario, input data is often fed in `batches'. We can however imagine that, if we had the ability to increase the number of steps at will (i.e., make $N$ larger while keeping constant the amount of data), we could formalize steps small enough to make the two processes comparable. 

Secondly, language is often seen as something that the learner has acquired and gained knowledge of. We want to bring back in the framework the role of the linguist-observer, that builds an abstraction over the linguistic behavior of the speaker (\S\ref{language}).
As the actual knowledge acquired by the speaker is undetectable and only explainable metalinguistically, in a way that is not viable with neural networks (i.e., we cannot ask NLMs what they know about linguistic regularities), we must take into account the fact that we are always analyzing both the linguistic input received by the speaker and the output produced as an effect of the acquisition process through analytical categories that are created and used by the linguist-observer. In other words, $\Lambda$ is not a property of the speaker, but rather a function operated by the linguist-observer. It does not evolve per se during the acquisition process, but rather it helps us detect and characterize the evolution of the speaker's abilities.

%\noindent{}describes both human and artificial acquisition processes, and its components have been central in the linguistic debate.

\subsection{The process of acquisition}
\label{steps}

All the elements of Equation~\ref{eq1} ideally change throughout time as the acquisition process unfolds.

The input $I$ to which the learner is exposed, in a real-life scenario, changes continuously. We can therefore define $I = (\iota_0, \iota_1, \cdots, \iota_N)$, where $\iota_i$ is the collection of input data to which the learner has been exposed to in-between $a_{i}$ and $a_{i+1}$. Again ideally, with $N$ large enough, each $\iota_i$ could even correspond to a single sentence.
The computational complexity also co-evolves with the acquisition function, as linguistic knowledge gets incorporated into it. In the human case, the initial state is unobservable and in the artificial scenario it is often not interesting as initialization of neural models is random. At step $i$, instead, the computational mechanism that has incorporated knowledge up to step $i-1$ is exposed to $\iota_i$. For these reasons, we define $C = (c_\emptyset, c_0,\cdots, c_{N-1})$.
As an effect, $\Lambda$ identifies different subsets $\lambda_0, \lambda_1, \cdots, \lambda_N$ throughout the acquisition process, namely $\Lambda = \bigcup\limits_{i=0}^N\lambda_i$

Each step of the broader process $A$ can be therefore defined as:
\begin{equation}
\begin{cases}
a_0 : \iota_0 \times c_\emptyset \mapsto \lambda_0 \\
a_i : \iota_i \times c_{i-1} \mapsto \lambda_i
\end{cases}
\end{equation}

\subsection{How do we observe \textit{learned} language?}
\label{language}

The notion of language that we introduced incorporates that of grammar, namely the analytical categories that we superimpose on the linguistic stream in order to analyze it and its unfolding over time. We do not test language as a cognitive state of the speaker: we intend it instead a set of categories that the observer (i.e., the linguist) considers relevant to the description of the linguistic stream produced by the (artificial) speaker. 
There exists, therefore, a striking difference between the linguistic stream (either the input perceived or the output produced by the speaker) and its representation through the lens provided by \textit{language}. 

%Let's focus therefore a little more on $\Lambda$ or, more specifically, each $\lambda_i$ and how we're able to observe it. 
If we wanted to be more precise with the notation, we should acknowledge the fact that language, i.e. $\Lambda$, as we mean it is actually a function by itself, that takes as input some linguistic stream (some observable data) and returns a representation of it. We could therefore rewrite the definition of $a_i$ as $a_i: \iota_i \times c_{i-1} \mapsto \Lambda(o_i)$ where $o_i$ is the linguistic stream produced by the speaker as a result of acquisition step $a_i$.

%As we defined earlier, $\lambda_i$ stands for the language that the (artificial) speaker has \textit{learned} by step $i$. From the linguistic perspective, \textit{language} refers to a deeper concept than the linguistic material uttered (or perceived) by the the speaker.

As we are interested in the categories that are acquired by the speaker and deployed during language comprehension and production, defining $\lambda_{o_i} = \Lambda(o_i)$ allows us to apply the same transformation on the input $\iota_i$ to which the speaker is exposed, thus obtaining $\lambda_{\iota_i}$ that is comparable to $\lambda_{o_i}$ in terms of linguistic categories.

Sticking to the constructionist perspective while trying to make the fewest possible assumptions on the actual content of linguistic knowledge, we hypothesize language as made up of a network of structures that are supposed to approximate constructions.
As constructions are form-meaning pairs, the notion of grammar incorporates that of a meaning space spanning beyond the lexical level. This can be easily implemented by extending the notion of vector space models that has been extensively explored and used in distributional semantics~\citep{lenci2008distributional,erk2012vector,lenci2018distributional}.
This represents a major difference with nativist approaches and the standard evaluation framework: meaning cannot be factored out of grammar effects and the acquisitional framework must account for its role in the process.
If we had to formalize the content of any $\lambda_i$, therefore, we could expand it as $\lambda_i = \{ (\kappa, \Vec{\kappa}) \mid \kappa \text{ is a construction wrt. some linguistic stream} \}$

\begin{figure}[t]
    \centering
    \includegraphics[width=0.5\textwidth]{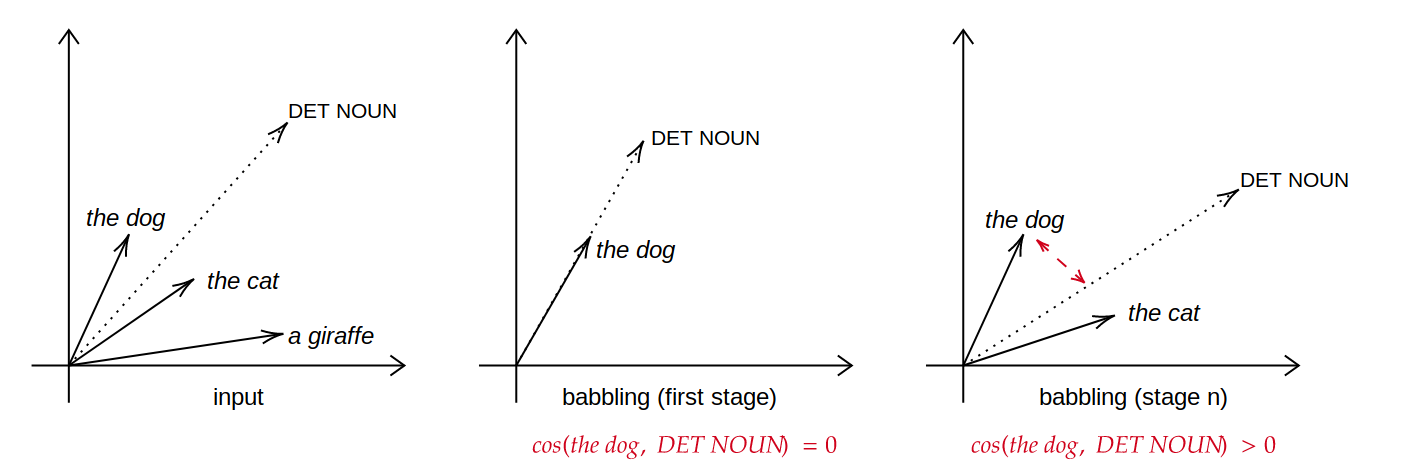}
    \caption{Let's assume that $\Lambda$ contains both contructions \textit{DET NOUN} and \textit{the dog}, with the latter being a lexicalized instance of the former. At different steps during acquisition, the two constructions can assume different meanings and be therefore associated with different distributional vectors. A distributional vector condenses in fact information about co-occurrences between linguistic items in a given piece of text. In the figure, we see that a cluster of vectors gather around \textit{DET NOUN} in the constructicon built from the input data (leftmost panel). This means that a variety of lexicalized instances exist for the construction \textit{DET NOUN}. During learning, the constructicons built from generated output show different distributions for the construction \textit{DET NOUN}. In the central panel, the cosine distance between \textit{DET NOUN} and \textit{the dog} is 0, meaning that their distributional contexts (i.e., their co-occurrences) perfectly overlap. In the rightmost panel instead, the distance between the two vectors has increased as another lexicalized instance (i.e., \textit{the cat}) is being produced. In this scenario, the contexts where \textit{DET NOUN} appears do not perfectly overlap with those where \textit{the dog} appears.}
    \label{fig:distributional shift}
\end{figure}

Unpacking this, we are saying that each obtained constructicon $\lambda_i$ is a network of structures. These can be more or less lexicalized, with their abstractness being a proxy for linking the structures in the network as we will explain in the next paragraph. Each construction is associated with a distributional vector (Figure~\ref{fig:distributional shift}), which represent its meaning.

\subsection{Desiderata: the structure induced by $\Lambda$}

We defined $\Lambda$ as a function that takes as input a linguistic stream $\tau$ and returns a \textit{constructicon} $\lambda_\tau$: a structured repository of form-meaning pairs.
In order to define and explore the internal structure of the constructicon, we introduce a few auxiliary functions and definitions:\\
\textbf{(i):} having meaning defined as a distributional space allows for distance computation $d(\kappa_i, \kappa_j)$ with $d: \Lambda \times \Lambda \mapsto [0, 1]$. $d(\cdot, \cdot)$ is a metric function that computes the distance between two meaning vectors. Usually, $d(\kappa_i, \kappa_j) = 1-cos(\Vec{\kappa_i}, \Vec{\kappa_j})$, where $cos(\Vec{\kappa_i}, \Vec{\kappa_j})$ is the cosine similarity between the two vectors associated to $\kappa_i$ and $\kappa_j$;\\
\textbf{(ii):} constructions bearing different abstraction levels are linked in the network. In order to navigate the network we introduce the function $c(\kappa_i, \kappa_j)$ with $c: \Lambda \times \Lambda \mapsto \{0, 1\}$ being a boolean function that computes whether two constructions constitute an \textit{abstraction chain}. For instance, $\kappa_i = \text{nsubj, GIVE, iobj, dobj}$ and $\kappa_j = \text{nsubj, root, iobj, dobj}$ form a chain with $\kappa_i$ being a partially lexicalized (hence, less abstract) instance of $\kappa_j$.

\subsection{Use scenarios}
\subsubsection{Individual acquisition over time}

The framework can be used to observe how the acquisition process unfolds over time. 
We can in fact set a number of steps $n$ and observe: (i) how the shape of grammar changes over the course of learning, comparing the various steps, as in:  $\Lambda(o_1) \sim \Lambda_(o_2) \sim \cdots \sim \Lambda_(o_n)$, (ii) how the grammar of the input can be compared to that acquired by the speaker, as in: $\Lambda(I) \sim \Lambda(o_n)$.
Given a subset $K \subseteq \Lambda(I)$\footnote{Actually, we have to make sure that $K \subseteq \Lambda(I) \cap \lambda_0 \cap \lambda_1 \cap \cdots \cap \lambda_n$} of interesting constructions, we can observe their behaviour over the learning process.

A popular constructionist hypothesis~\citep{goldberg2006constructions}, for example, states that the meaning of a construction (e.g., the ditransitive pattern \textit{Subj V Obj Obj2}), and therefore its productivity, emerges from the association with specific lexical items in the input received by the learner (e.g., \textit{give} in the case of the ditransitive): part of the lexical meaning remains attached to the meaning of the syntactic pattern, and therefore its distributional properties with it.
Let's assume that the speaker has acquired some construction $\kappa$ (e.g., the ditransitive construction). Once they're able to use it in a productive and creative way (i.e., in a more varied contexts than the \textit{give} contexts the construction is strongly associated with in the input), we can use the proposed framework to check whether the distributional meaning of two constructions $\kappa_i, \kappa_j \in \Lambda(I)$ with $c(\kappa_i, \kappa_j) = 1$ (i.e., with $\kappa_i$ being a less abstract instance of $\kappa_j$) influences the learnability of $\kappa_j$ as an independent construction.

The notion of \textit{abstraction chain} introduced before helps us testing this hypothesis as we can check the behaviour of the chain $(\kappa_i, \kappa_j)$ at each timestep.
We can denote $\kappa_i^{\lambda_k}$ the construction $\kappa_i \in \lambda_k$ and similarly $\kappa_j^{\lambda_k}$ the construction $\kappa_j \in \lambda_k$, through distributional analysis we can capture how the contexts in which $\kappa_i$ and $\kappa_j$ vary, and whether this variation is associated with grammatical generalization. We expect, in fact, $d(\kappa_i, \kappa_j)$ to increase during acquisition:
\begin{equation}
    d(\kappa_i^{\lambda_a}, \kappa_j^{\lambda_a}) \leq  d(\kappa_i^{\lambda_b}, \kappa_j^{\lambda_b}) \;\; \forall \: a, b \mid a \leq b
\end{equation}

If $\kappa_j$ is produced in contexts that do not perfectly overlap with those where $\kappa_i$ is produced, this indicates that the speaker has gained a productive use of construction $\kappa_j$, which is recognized as an independent construction from $\kappa_i$. If conversely their distance decreases during acquisition, we might deduce that the speaker has recognized $\kappa_j$ as unnecessary by restricting its application cases to those of $\kappa_i$.

\subsubsection{Language as the expression of a population of speakers}

We are often interested in defining grammar in terms of what can be considered shared linguistic knowledge among the speakers.
A core aspect of construction grammar is in fact conceiving language primarily as a social and external phenomenon, as opposed to nativist theories that focus on its inner nature.
By means of the framework, we can analyze grammar as an abstraction over the linguistic productions of a population of $P$ speakers $\Pi = (\sigma_1, \sigma_2, \cdots, \sigma_P)$. We can define the grammatical conventions deployed by the community $\Pi$ as $\Lambda_\Pi = (\lambda_{\sigma_1}, \lambda_{\sigma_2}, \cdots, \lambda_{\sigma_P})$.
This allows for modelling variation between the acquisition process of the different speakers. Speaker $\sigma_i$ might be exposed to a unique series of input material $\iota_0^{\sigma_i}, \cdots, \iota_N^{\sigma_i}$ that does not necessarily coincide with that of speaker $\sigma_j$.

In this setting, we can for instance investigate what is learned \textit{no-matter-the-input}, and what is instead specific or idiosyncratic for each speaker.
We can define:
\begin{equation}
    G_{\geq p} = \left\{ \kappa \mid \sum\limits_{i=0}^P X(\kappa, \sigma_i) \geq p \right\} 
\end{equation}
\noindent{}as the set of constructions that we can observe in the linguistic productions of $p$ or more speakers. With:
\begin{equation}
    X(\kappa_i, \sigma_j) = 
    \begin{cases}
    1 & \text{if $\kappa \in \Lambda^{\sigma_j}$} \\
    0 & \text{otherwise}
    \end{cases}
\end{equation}
\noindent{}being an auxiliary function that evaluates to $1$ if the construction $\kappa$ appears in the production of speaker $\sigma_j$ and $0$ otherwise (this just helps us count how many speakers use construction $\kappa$ productively).
$G_P$ would for instance be the set of constructions shared by all speakers in a population, and could be therefore identified as the set of \textit{core} constructions in $\Lambda^\Pi$. When, instead, $p \ll P$, we are observing constructions that are not shared by a significant amount of speakers in the population, and their use can therefore depend on specific input instances or tendencies in subgroups of speakers.
Following the same logic we can of course also just define $G_{(\sigma_i, \sigma_j)}$ as the constructions that are common to the two speakers $\sigma_i$ and $\sigma_j$.
By means of $G$, we can define $\widetilde{\Lambda_G}$ as an approximation of the function $\Lambda$, which only uses the categories that are retained in $G$. $\widetilde{\Lambda_{G_{\geq P}}}$ would for instance be a function that considers only linguistic knowledge shared by the entire population $\Pi$, while $\widetilde{\Lambda_{\sigma_i}}$ would be restricted to the constructicon $\lambda_{\sigma_i}$. 
Considering speakers $\sigma_i$ and $\sigma_j$, with their respective produced linguistic outputs $O_{\sigma_i}$ and $O_{\sigma_j}$, we can produce and compare $\widetilde{\Lambda_{G_{\sigma_i}}}(O_{\sigma_j})$ and $\widetilde{\Lambda_{G_{\sigma_j}}}(O_{\sigma_i})$: respectively, what speaker $\sigma_i$ is able to retrieve from $O_{\sigma_j}$ and what speaker $\sigma_j$ is able to retrieve from $O_{\sigma_i}$.

%With this approach we can easily build populations of speakers that acquire language on inputs varying on a number of sociolinguistic parameters: the input can be more or less lexically varied, the speaker can be exposed to increasingly complex linguistic materials or not, the distribution of linguistic categories in the input could be more or less skewed... We can monitor how variations in the input received by speakers gets reflected in their linguistic productions.
The fact that speakers use the same constructions $\kappa$ to build their linguistic productions does not of course ensure that the corresponding meanings $\vec{\kappa}$ coincide.\footnote{This makes sure that $G_{\sigma_i}$ does not coincide with $\lambda_{\sigma_i}$} Different speakers, depending on the input they have been exposed to, and to the partial randomness attributed to computational mechanisms, could associate different meaning spaces to the same construction. 
Given two speakers $\sigma_i$ and $\sigma_j$, and a sentence $s$, we could therefore compare the portions of $\lambda_{\sigma_i}$ and $\lambda_{\sigma_j}$ meaning spaces that are activated to linguistically (de)compose the sentence $s$.

\section{Exploratory experiments}
\label{sec:experiments}

\begin{figure}[t]
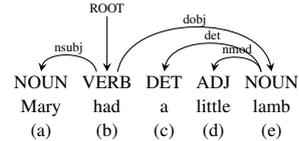

	\centering
	\begin{scriptsize}
	\begin{dependency}[theme = simple]
		\begin{deptext}[column sep=0.3em, row sep=.1ex]
			NOUN \& VERB \& DET \& ADJ  \& NOUN \\
			Mary \& had  \& a  \& little \& lamb  \\
			 (a)  \& (b)  \& (c) \& (d)   \& (e) \\   
		\end{deptext}
		\deproot[edge unit distance = 2ex]{2}{ROOT}
		\depedge{2}{1}{nsubj}
		\depedge{2}{5}{dobj}
		\depedge{5}{3}{det}
		\depedge{5}{4}{nmod}
	\end{dependency}
	\caption{The dependency representation of the sentence \textit{Mary had a little lamb}, annotated with morpho-syntactic and syntactic information. In this structure, we can identify the following \textit{catenae}: a, b, c, d, e, ab, abce, abde, abcde, abe, bce, bde, be, ce, de, cde. Other possibilities would have been \textit{strings} (e.g., a, ab, abc, ... b, bc, ...e) or \textit{constituents} (i.e., a, abcde, c, d, cde).}
	\label{fig:mary-had-a-little-lamb}
	\end{scriptsize}
\end{figure}

In order to explore the potential applications of the framework described in \S\ref{sec:formalization}, we built a simple instance using the CHILDES corpus~\citep{macwhinney2000childes} as input data $I$ and a vanilla character-based LSTM~\citep{hochreiter1997long} as computational mechanisms $C$. With this simple setting, we explored two aspects: (i) we tested whether distributional similarities in $\lambda_I$ would influence the acquisition of constructicons $\lambda_1, \cdots, \lambda_n$, and (ii) we tried to describe grammar as it emerges from a population of speakers.
Constructions were approximated through \textit{catenae}~\citep{osborne2012catenae}: subtrees extracted from a dependency parsing syntactic representation (see Figure~\ref{fig:mary-had-a-little-lamb}).

%Implementation-wise, structures are automatically extracted from parsed text in form of \textit{catenae}. As not every possible pattern is retained, the created repository is then reduced so a subset of meaningful structures that are supposed to be the ones that actually participate in linguistic knowledge. The reduction process is done by means of  standard association measures.

%Having highlighted the important methodological issue related to the use of computational modelling for linguistic research, we now introduce our proposal for a different methodology to investigate the acquisition of grammar by NLMs in a constructionist perspective.
%In the first part, we will deal with theoretical aspects, reviewing the recent literature in the light of various linguistic assumptions and showing how a model statement would clarify such assumptions. In the second part, we will propose a case study consisting of two specific experiments, and show how explicit model statement helps analysing and interpreting the results of those experiments.

\subsection{Abstracting grammar over training}

We first replicate an analysis presented in~\citet{pannitto2020recurrent}, where a character-based LSTM was trained on CHILDES corpus. The authors fixed 7 steps during the LSTM's acquisition process, each after 5 epochs of training.
In our formalization, this equates to 7 constructicons $\lambda_1$ to $\lambda_7$. The distributional space for each $\lambda_i$ is obtained by counting co-occurrences between constructions within the same sentence.
We can then consider abstraction chains $(\kappa_i, \kappa_j)$ in $I$ (i.e., in $\Lambda(\text{CHILDES})$ and computed $d(\kappa^{\lambda_7}_i, \kappa^{\lambda_7}_j) - d(\kappa^{\lambda_1}_i, \kappa^{\lambda_1}_j)$ for each abstraction chain, namely the difference in cosine similarity between step 7 and step 1. Grouping all chains by $\kappa_i$ and $\kappa_j$, it is possible to compute the average distributional shift as shown in Table~\ref{tab:dist-shift} (i.e., for each $\kappa_i$ to its more abstract instances and for each $\kappa_j$ to its more concrete instances).

\begin{table}[t]
\begin{scriptsize}
\begin{tabular}{p{0.1\textwidth}cc|p{0.1\textwidth}cc}
$\kappa_i$ & \textbf{shift} & \textbf{cosine}  & $\kappa_j$ &  \textbf{shift} & \textbf{cosine}  \\ \hline
\tiny{@nsubj @root so} & 0.18 & 0.43 & \tiny{more @root} & 0.2 & 0.21 \\
\tiny{@nsubj only @root} & 0.18 & 0.41 & \tiny{\_AUX know @obj} & 0.19 & 0.66 \\
\tiny{what @root @obj} & 0.18 & 0.39 & \tiny{@advmod tell} & 0.17 & 0.64 \\
\tiny{what @advmod \_VERB} & 0.16 & 0.19 & \tiny{@aux know @obj} & 0.16 & 0.71 \\
\tiny{only @root} & 0.16 & 0.38 & \tiny{@advmod can \_VERB} & 0.15 & 0.76 \\
\tiny{more @root} & 0.16 & 0.23 & \tiny{know @obj} & 0.15 & 0.62 \\
\tiny{@root it @xcomp} & 0.15 & 0.61 & \tiny{a \_NOUN} & 0.13 & 0.52 \\
\tiny{@det minute} & 0.15 & 0.25 & \tiny{might @root} & 0.13 & 0.70 \\
\tiny{\_PRON only @root} & 0.15 & 0.53 & \tiny{\_PRON @root n't} & 0.12 & 0.53 \\
\tiny{\_VERB \_DET minute} & 0.15 & 0.33 & \tiny{@root that \_VERB} & 0.12 & 0.65 \\
\tiny{\_PRON @root so} & 0.14 & 0.54 & \tiny{\_VERB 'll @ccomp} & 0.12 & 0.71 \\
\tiny{\_DET minute} & 0.134 & 0.33 & \tiny{\_VERB me @obl} & 0.12 & 0.76 \\
\end{tabular}
\caption{Constructions with highest average shifts.}
\label{tab:dist-shift}
\end{scriptsize}
\end{table}

\begin{figure}[t]
	\centering
	\includegraphics[width=0.4\textwidth]{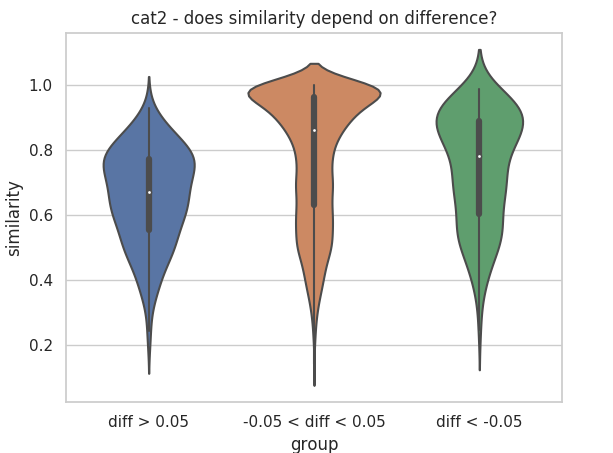}
	\caption{Distribution of average cosine similarities for the three groups of $kappa_j$, showing low, intermediate and high average shifts respectively.}
	\label{fig:similarity-difference-2}
\end{figure}

Three bins are considered, based on average distributional shift: the hypothesis is that constructions that underwent the highest shifts during training are those showing intermediate levels of similarities in the input distributional space. Indeed, chains with very high input similarities are unlikely to exhibit abstraction: according to constructionist intuition, their distributional similarity means that the construction that is part of the \textit{Constructicon} is the least schematic one, and there is no need for the more schematic (and therefore, \textit{abstract}) category to be created. Low similarity pairs, on the other hand, may simply contain unrelated constructions. The three groups show different distributions\footnote{A Kruskall-Wallis one-way ANOVA was performed and resulted in significant values.} as shown in Figure~\ref{fig:similarity-difference-2}.

\subsection{A population of artificial speakers}

Following on~\citet{pannitto2020recurrent} experiments, we consider a population of 10 speakers modeled with 10 vanilla character-based LSTMs trained on random samples of the CHILDES corpus (each containing 1 million words).

With this setting, we try to identify the locus of variation among different speakers, under the assumption that some `core' constructions \textit{must} be shared by all individuals, while others are less important to successful communication.

We restrict the analysis to the constructions to which all 10 speakers have been exposed to through their input (11051 constructions) and create $G_{10}$ as the set of \textit{core constructions} and $G_{\leq 5}$ as the set of \textit{periphery constructions}, i.e. the ones shared by half of the speakers or less.
Being trained on random samples taken from the same distribution, the speakers share most of the constructions (9086 out of 11051). However, we expect these numbers to change significantly when the input language varies along more refined sociolinguistic axes.

\begin{figure}[t]
	\centering
	\includegraphics[width=0.4\textwidth]{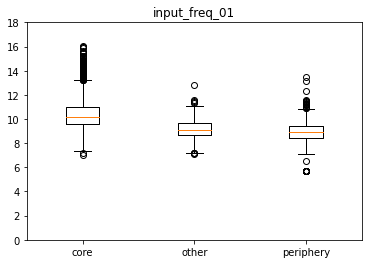}
	\caption{Difference in input frequency between the three groups of constructions: \textit{core} as the ones shared by all speakers, \textit{periphery} as the ones shared by half of the speakers or less, and \textit{other} as the remaining ones.}
	\label{fig:speaker01}
\end{figure}

We also checked, for all speakers, whether the constructions of the \textit{core} group and the constructions of the \textit{periphery} group had significantly different frequencies in the input given to each speaker. As shown in Figure~\ref{fig:speaker01}, the difference between the three groups are significant despite not appearing as striking as one would expect.

Lastly, we explored the input through $\widetilde{\Lambda_{G_{10}}}$ and  $\widetilde{\Lambda_{G_{\leq 5}}}$, as shown in Table~\ref{tab:translated-sentence}: both representations (the one obtained through \textit{core} constructions and the one obtained through \textit{periphery} constructions) highlight meaningful patterns in the sentence, but only the former can be considered a grammatical representation shared by the population.

\begin{table}[t]
\centering
\begin{scriptsize}
\begin{tabular}{l|l|l}
\textbf{corpus}  & \textbf{Core} & \textbf{Periphery} \\ \hline
does    & AUX    &  -      \\
n't     & n't    &  -     \\
that    & that   &  PRON       \\
seem    & @root  &  VERB      \\
kind    & -      &  ADV     \\
of      & -      &  -     \\
silly   & ADV    &  ADV
\end{tabular}%
\caption{A sentence (left column) as it would appear if we restricted to only \textit{core} (middle column) or \textit{periphery} (right column) constructions.}
\label{tab:translated-sentence}
\end{scriptsize}
\end{table}

\section{Concluding remarks}

%What has all of this to do with Cognitive Science? 
The nature of linguistic representations is a core issue in linguistic theories of language development. We feel this aspect has been overlooked in the NLMs literature and propose an approach that brings back theoretical insights into the picture.
We commit here to the UB constructionist framework, not as an ideal model of human language acquisition, but rather as a set of tools and categories that suffice to explain NLMs' generated language.

%Nativist theories represented and still represent a fundamental approach to language sciences, having greatly contributed to the understanding of aspects of both language and the mind. However, their legacy has often spread perhaps outside of the boundaries and assumptions that were posited by Chomsky and his entourage. As \cite{christiansen2016creating} note, isolating the study of language from considerations regarding processing, acquisition and evolution has affected the way researchers have approached the observation of linguistic phenomena outside of the UG theory strictu sensu.
Since learning a language largely overlaps with learning to process the input, there must be a relation between processing biases relating to certain types of constructions and the distribution of those constructions in the linguistic input~\citep{christiansen2016creating}. As experience grounds linguistic knowledge, distributional properties become a key aspect to determine the content of linguistic representations. In this framework, language is not considered as an autonomous cognitive system. Rather, the acquisition of grammar is regarded as any other conceptualization process and knowledge of language emerges from use.
%In construction grammars, constructions are seen as pairings of form and meaning, and meaning is intended as all the conventionalized aspects of a construction's function~\citep{Croft2004-cognitive}. 

To conclude, the observation of NLMs linguistic abilities would benefit from a constructionist approach.
The evaluation can take place at multiple levels and includes properties of the situation described by the linguistic signal, but also properties of the linguistic signal itself. The UB framework may in fact provide useful categories to analyze the statistical properties of artificial language learners, and most importantly allows us to examine the semantic and the syntactic layers in parallel, both in the input received by the learner and in the stochastic output it generates. 
\bibliography{anthology,custom}
\bibliographystyle{acl_natbib}

%\appendix

%\section{Example Appendix}
%\label{sec:appendix}

%This is an appendix.

\end{document}